\documentclass[11pt,a4paper]{article}

\usepackage[utf8]{inputenc}
\usepackage[T1]{fontenc}
\usepackage[margin=2.5cm]{geometry}
\usepackage{graphicx}
\usepackage{hyperref}
\usepackage{authblk}
\usepackage{xcolor}
\usepackage{amsmath, amssymb}
\usepackage{bm}
\usepackage{mathtools}
\usepackage{siunitx}
\usepackage{subcaption}
\usepackage{enumitem}

\graphicspath{{fig/}}

\hypersetup{
    colorlinks=true,
    linkcolor=blue,
    citecolor=blue,
    urlcolor=cyan,
}

\DeclareMathOperator*{\argmax}{arg\,max}

\newcommand{\diff}[1][]{\mathrm{d}#1}

\newcommand{\diffrac}[3][]{\frac{\diff^{#1}#2}{\diff#3^{#1}}}

\renewcommand{\vec}[1]{\bm{#1}}
\newcommand{\norm}[1]{\left\lVert#1\right\rVert}

\newcommand{\setEx}[1]{\ensuremath{\left\{#1\right\}}}

\title{\textbf{Controlling Fish Schools via Reinforcement Learning of Virtual Fish Movement}}

\author[1]{Yusuke Nishii}
\author[1,2]{Hiroaki Kawashima\thanks{Present address: Graduate School of Information Science, University of Hyogo, Kobe, Japan. Corresponding author: kawashima@gsis.u-hyogo.ac.jp}}

\affil[1]{Faculty of Engineering, Kyoto University, Kyoto, Japan}
\affil[2]{Graduate School of Informatics, Kyoto University, Kyoto, Japan}

\date{}

\begin{document}

\maketitle

\begin{center}
    \begin{minipage}{0.9\textwidth}
        \small
        \itshape
        \textbf{Note:} This report presents an English translation of the original Japanese bachelor's thesis submitted by Yusuke Nishii in 2018 to the Undergraduate School of Electrical and Electronic Engineering, Faculty of Engineering, Kyoto University, under the supervision of Hiroaki Kawashima. The research was conducted between 2017 and 2018 while the authors were affiliated with Kyoto University.
        This translation includes minor revisions for clarity. The original thesis was completed on February 9, 2018, and its content reflects the state of research at that time. The translation aims to preserve the meaning and structure of the original work while making it accessible to a broader audience. 
        The literature review and the scientific context of this work reflect the state of the field at the time of the original submission date.
    \end{minipage}
\end{center}

\vspace{1em}

\begin{abstract}
This study investigates a method to guide and control fish schools using virtual fish trained with reinforcement learning. We utilize 2D virtual fish displayed on a screen to overcome technical challenges such as durability and movement constraints inherent in physical robotic agents. To address the lack of detailed behavioral models for real fish, we adopt a model-free reinforcement learning approach. 
First, simulation results show that reinforcement learning can acquire effective movement policies even when simulated real fish frequently ignore the virtual stimulus. Second, real-world experiments with live fish confirm that the learned policy successfully guides fish schools toward specified target directions. Statistical analysis reveals that the proposed method significantly outperforms baseline conditions, including the absence of stimulus and a heuristic ``stay-at-edge'' strategy. 
This study provides an early demonstration of how reinforcement learning can be used to influence collective animal behavior through artificial agents.
\end{abstract}

\section{Introduction}
\label{sec:introduction}

\subsection{Research Background}
The long-term objective of this research is to guide and control the movement of fish schools using externally controlled robotic agents. Potential applications range from choreographing fish school trajectories for aquarium exhibitions to navigating schools in aquaculture settings.

In such practical scenarios, the target fish schools are expected to be relatively large, making it impractical to provide stimuli to every individual within the group. Consequently, it becomes necessary to control the entire school by stimulating only a limited number of individuals. To achieve this, it is crucial to provide stimuli tailored to the real-time state of the school, for instance, by identifying which individuals exert the strongest influence on the collective motion and determining how they should move to guide the group toward a desired objective. This suggests that feedback control, which provides appropriate stimuli based on the evolving state of the school, is indispensable.

A primary approach to such control involves the use of motion prediction models for fish schools. These models would allow for predicting the school's response to specific stimuli and pre-evaluating control strategies against desired targets. Regarding motion models, previous studies, such as the one by Couzin et al.~\cite{Couzin:2002}, have successfully simulated natural schooling behaviors and contributed significantly to elucidating collective mechanisms. However, at present, these models have not yet reached the level of accuracy required to predict the real-time movements of actual fish schools.

\subsection{Research Objectives}

Potential stimuli for influencing fish behavior include predators, food, and artificial conspecifics. However, the use of predators is expected to cause significant stress to the target fish, while the continuous provision of food can lead to health and environmental issues. Therefore, this study focuses exclusively on the use of artificial conspecifics as stimuli.

To address the challenge of not having a predictive model for the target school's movement, we propose an approach that acquires movement policies through reinforcement learning (RL)~\cite{Sutton:1998}, with the artificial conspecific serving as the RL agent. RL is a learning framework in which an agent observes the state of the environment, determines an action based on that state, and receives a reward as an evaluation of that action to acquire a near-optimal policy. This method offers the distinct advantage of not requiring an explicit model of the environment (the target schooling behavior).

While the long-term goal is to implement these artificial conspecifics as physical robots, controlling the precise motion of the robots themselves remains a significant challenge. To circumvent this, we utilize virtual fish displayed on a screen. Virtual fish not only allow for easier motion control but also offer superior durability compared to mechanical robots. Since RL requires virtual fish to interact with the school over extended periods to acquire an effective policy, the durability of virtual agents is highly suitable for the objectives of this study.

Based on the above, this research aims to investigate a method for controlling fish schools by applying RL to the movements of virtual fish. For simplicity, the scale of the fish school is limited to a small number of individuals.

\subsection{Related Work}

Various models have been proposed to describe collective behavior in nature. Notable examples include the Kuramoto model~\cite{Kuramoto:1975,Rodrigues:2016}, originally developed to explain the synchronization of firefly flashing; the Boids model~\cite{Reynolds:1987}, which simulates the flocking behavior of birds; and the model by Couzin et al.~\cite{Couzin:2002}, which characterizes the collective movements of fish schools.

Several studies have investigated schooling behavior by providing external stimuli to fish. For instance, Swain et al. observed the reactions of fish to predator-like stimuli~\cite{Swain:2012}. Kopman et al. developed a robotic conspecific that adjusts its tail-fin motion based on the real-time positions of live fish to observe their behavioral changes~\cite{Kopman:2013}. Furthermore, research has also focused on using visual stimuli to influence behavior. Kawashima et al. utilized a camera-display system to present virtual fish to a school, confirming that the real fish synchronized their movements with the reciprocating motion of the virtual fish~\cite{Kawashima:2014}.

A fundamental biological reaction of fish to visual stimuli is the optomotor response. This is considered a reflexive behavior in which fish follow a moving object or pattern. A well-known experimental demonstration involves placing a cylinder with vertical black-and-white stripes around a water tank; as the cylinder rotates, the fish swim along with the stripes~\cite{Arnold:1974}. Such reflexive behaviors suggest the significant potential for controlling the movement of fish schools through precisely designed visual stimulation.

\subsection{Structure of the Report}

The remainder of this report is organized as follows. 
Section~\ref{sec:framework} describes the problem formulation of this study and the configuration of the interaction system between the virtual fish and the real fish. 
Section~\ref{sec:methodology} explains how the interaction system is modeled as a Markov Decision Process (MDP), a standard framework for representing agent-environment interactions in RL, and details the Q-learning algorithm employed in this study. 
Section~\ref{sec:simulation} presents a preliminary evaluation using simulations to verify whether RL remains effective even when simulated real fish frequently ignore the virtual fish. 
Section~\ref{sec:realworld} discusses the implementation of RL in a real-world environment and evaluates the performance of the acquired movement policies. 
Finally, Section~\ref{sec:conclusion} concludes this study and suggests directions for future work.

\section{Interaction System for Fish Schools and Virtual Fish}
\label{sec:framework}

\subsection{Problem Formulation}
For simplicity, this study considers a fish school consisting of a small number of individuals of the same species. The target species should be easy to obtain and maintain, and possess strong schooling tendencies. To satisfy these criteria, we use the Rummy-nose tetra ({\it Hemigrammus bleheri}\footnote{{\it Hemigrammus bleheri} has been reclassified as {\it Petitella bleheri} in 2020, while the original name is retained in this translation to maintain consistency with the 2018 source text.}).

Generally, the positions of fish in a tank vary freely in three-dimensional (3D) space. However, in this study, the experiments are conducted in a narrow environment to treat the fish positions as two-dimensional (2D) coordinates. The internal dimensions of the tank used for housing and experiments are \SI{34.0}{cm} in width, \SI{14.0}{cm} in length (front-to-back distance), and \SI{28.3}{cm} in height. During experiments, partition plates are used to restrict the swimming area to \SI{30.4}{cm} in width, \SI{5.0}{cm} in length, and \SI{12.5}{cm} in height.

\begin{figure}[tbp]
    \centering
    \includegraphics[width=0.8\linewidth]{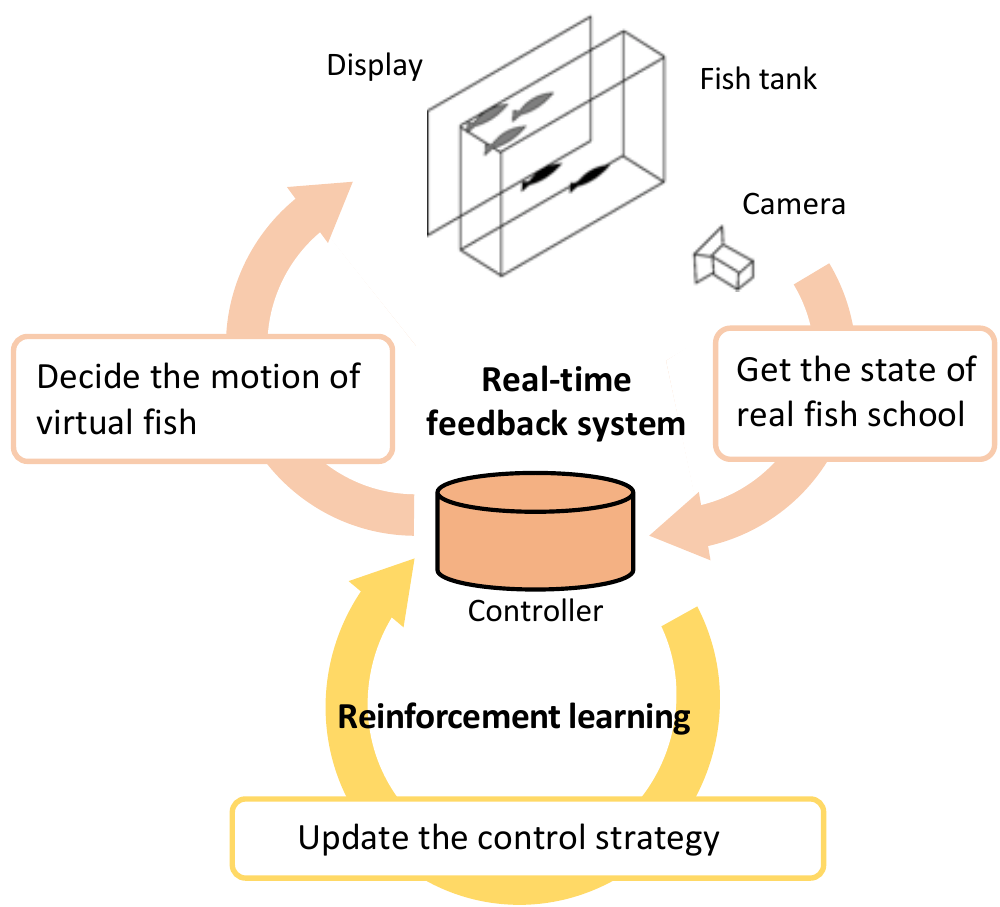}
    \caption{Overview of the interaction system between a fish school and virtual fish.}
    \label{fig:system-overview}
\end{figure}

A display is placed against the back of the tank to present virtual fish, and the interaction is observed using a camera positioned in front of the tank. A schematic diagram of this system is shown in Fig.~\ref{fig:system-overview}. Details regarding the presentation of virtual fish and the camera-based observation are provided in Sections~\ref{sec:display} and~\ref{sec:camera}, respectively.

\subsection{Rationale for Using Virtual Fish}
\label{sec:display}
First, we discuss the justification for using virtual fish on a screen. Fish schools are formed through reactions between individuals, categorized as follows:
\begin{description}
    \item[Attraction:] Individuals recognize conspecifics visually and move toward each other.
    \item[Alignment:] Individuals follow the movements of neighbors through the optomotor response.
    \item[Repulsion:] Individuals maintain distance from neighbors based on lateral line sensations to avoid collisions.
\end{description}
Among these, attraction leads to ``aggregation,'' where fish gather in close proximity, and alignment leads to the formation of a ``school,'' where they move in a unified direction. Therefore, we expect that the movement of the school can be controlled by inducing attraction and alignment, reactions primarily based on visual stimuli, using virtual fish. Indeed, it has been reported that when a school of three Rummy-nose tetras was presented with virtual fish performing periodic reciprocating motions via a camera-display system, the real fish tended to synchronize with the virtual ones~\cite{Kawashima:2014}. Thus, the use of virtual fish is considered valid.

\begin{figure}[tbp]
    \centering
    \includegraphics[width=0.6\linewidth]{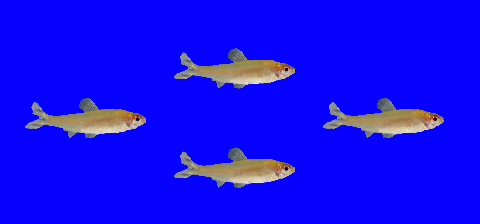}
    \caption{Image and arrangement of the virtual fish, including the background color.}
    \label{fig:imitated-texture}
\end{figure}

Next, we consider the design of the virtual fish. Since the front-to-back dimension of the swimming area is restricted, we place a display\footnote{
    ASUS MB168B (15.6-inch, resolution $1366 \times 768$, pixel pitch \SI{0.252}{mm}, response time \SI{11}{ms}, USB 3.0 interface).
} against the back of the tank. As previously mentioned, attraction is triggered by visual recognition of conspecifics. Therefore, it is effective to use virtual fish that resemble the target species in shape, size, and texture. Consequently, we use an image created by photographing real fish, as shown in Fig.~\ref{fig:imitated-texture}. The background color is kept uniform across the display.

\subsection{Camera-Based Measurement of Fish Schools}
\label{sec:camera}
\begin{figure}[tbp]
    \centering
    \includegraphics[width=0.8\linewidth]{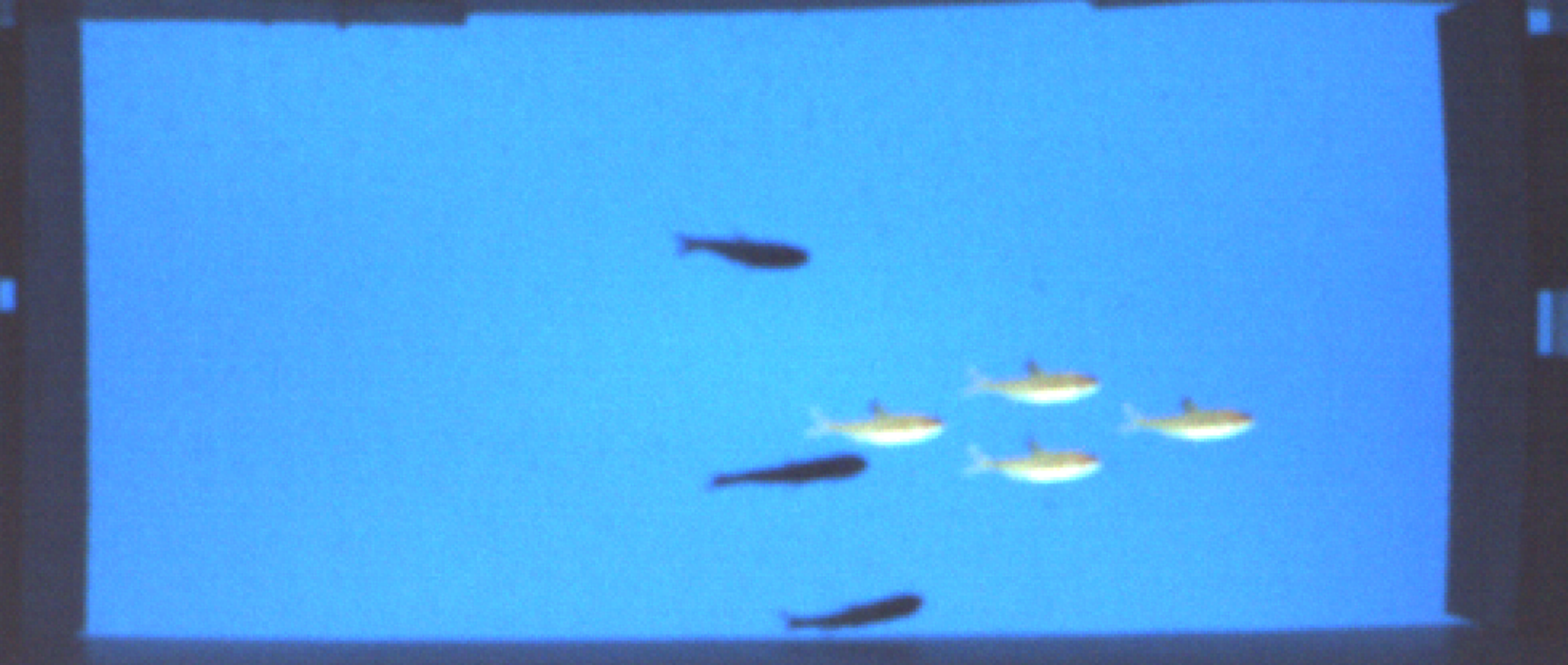}
    \caption{Example of a captured image. Due to the light from the display, the real fish appear dark as silhouettes.}
    \label{fig:observed-image}        
\end{figure}

A camera\footnote{
    FLIR Flea 3 FL3-U3-13E4C-C (resolution $1280 \times 1080$, color, max frame rate \SI{60}{fps}, USB 3.0 interface), lens: FUJINON HF9HA-1B (aperture F1.4--F16, focal length \SI{9}{mm}).
} is positioned in front of the tank to capture the interaction. The frame rate is set to \SI{10}{fps}, which is sufficient to capture the movements of the fish. An example of a captured image is shown in Fig.~\ref{fig:observed-image}. We define a ``viewport'' as a plane that collapses the front-to-back dimension within the restricted swimming area. Coordinates within the viewport are expressed as $(x_\mathrm{viewport}, y_\mathrm{viewport})$, normalized such that $x_\mathrm{viewport}, y_\mathrm{viewport} \in [0, 1]$.

The correspondence between the captured image coordinates $(x_\mathrm{camera}, y_\mathrm{camera})$ and the viewport coordinates $(x_\mathrm{viewport}, y_\mathrm{viewport})$ is approximately expressed as:
\begin{align}
    \label{eq:camera-to-viewport-x}
    x_\mathrm{viewport} &= a\,x_\mathrm{camera} + b,\\
    \label{eq:camera-to-viewport-y}
    y_\mathrm{viewport} &= c\,y_\mathrm{camera} + d.
\end{align}
The constants $a, b, c, d$ are obtained by manually selecting the four boundaries corresponding to the viewport in an image captured before the experiment.

Similarly, the relationship between the viewport coordinates and the display coordinates $(x_\mathrm{display}, y_\mathrm{display})$ is approximated as:
\begin{align}
    x_\mathrm{display} &= e\,x_\mathrm{viewport} + f,\\
    y_\mathrm{display} &= g\,y_\mathrm{viewport} + h.
\end{align}
To determine the constants $e, f, g, h$, two points with known display coordinates are shown before the experiment. Their positions in the captured image are manually identified and converted to viewport coordinates using Eqs.~\eqref{eq:camera-to-viewport-x} and~\eqref{eq:camera-to-viewport-y}. This allows us to calculate $e, f, g, h$ using the corresponding coordinate pairs in the display coordinate system and the viewport coordinate system.

Through this process, both the positions in the captured images and the positions on the display can be represented consistently within the viewport coordinate system.

\section{Reinforcement Learning of Virtual Fish Movement}
\label{sec:methodology}

\subsection{Markov Decision Process for Fish School Control}
\begin{figure}[tbp]
    \centering
    \includegraphics[width=0.6\linewidth]{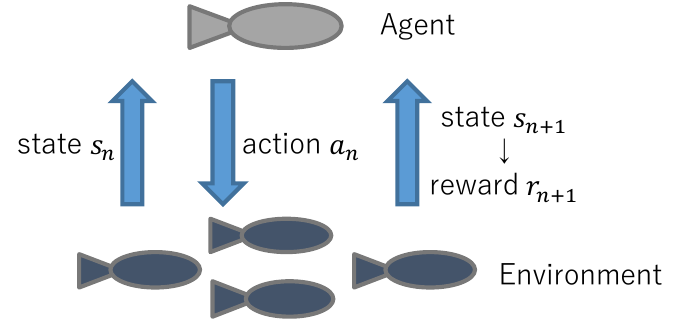}
    \caption{Interaction between the agent and the environment in reinforcement learning.}
    \label{fig:MDP}
\end{figure}

First, we outline the general framework of RL based on Sutton and Barto~\cite{Sutton:1998}. In RL, the entity that performs learning and decision-making is called the \textit{agent}, and the entity it interacts with is called the \textit{environment}. The agent and environment interact at each discrete time step $n = 0, 1, \dots$. Specifically, the agent observes a state $s_n \in \mathcal{S}$ (where $\mathcal{S}$ is the state space) from the environment and, based on this, selects an action $a_n \in \mathcal{A}$ (where $\mathcal{A}$ is the action space). By performing action $a_n$, the agent influences the environment, resulting in the observation of the next state $s_{n+1}$ and a corresponding reward $r_{n+1} \in \mathbb{R}$. This interaction is illustrated in Fig.~\ref{fig:MDP}.

When the system satisfies the Markov property, that is, when $s_{n+1}$ and $r_{n+1}$ are determined probabilistically only by $s_n$ and $a_n$, the model is called a Markov Decision Process (MDP). An MDP with finite sets $\mathcal{S}$ and $\mathcal{A}$ is known as a finite MDP. In many cases, RL aims to address situations that can be approximated as finite MDPs. The goal of the agent is to learn actions $a_n$ that maximize the long-term cumulative reward $r_{n+1}, r_{n+2}, \dots$, formulated as the maximization of the discounted return $R_n$ using a discount rate $\gamma \in [0, 1]$:
\begin{align}
    R_n = \sum_{k=0}^\infty \gamma^k r_{n+k+1}.
\end{align}

Here, we model the fish school control problem for RL. For simplicity, we focus on controlling the $x$-coordinate, the primary direction of movement, while ignoring the $y$-coordinate. The control objective is to guide the centroid of the real fish school toward the edges of the tank. The target direction is switched between the left and right edges at appropriate intervals. By assuming left-right symmetry, we consider that the policy learned to guide the school toward one edge is equally effective for the opposite direction.

The discrete time step $n$ corresponds to the time when the $(10n)$-th frame is captured by the camera; thus, the duration of each step is \SI{1}{s}. Continuous time is denoted by $t$, and the continuous time corresponding to step $n$ is denoted by $t_n$.

\begin{figure}[tbp]
    \centering
    \includegraphics[width=\linewidth]{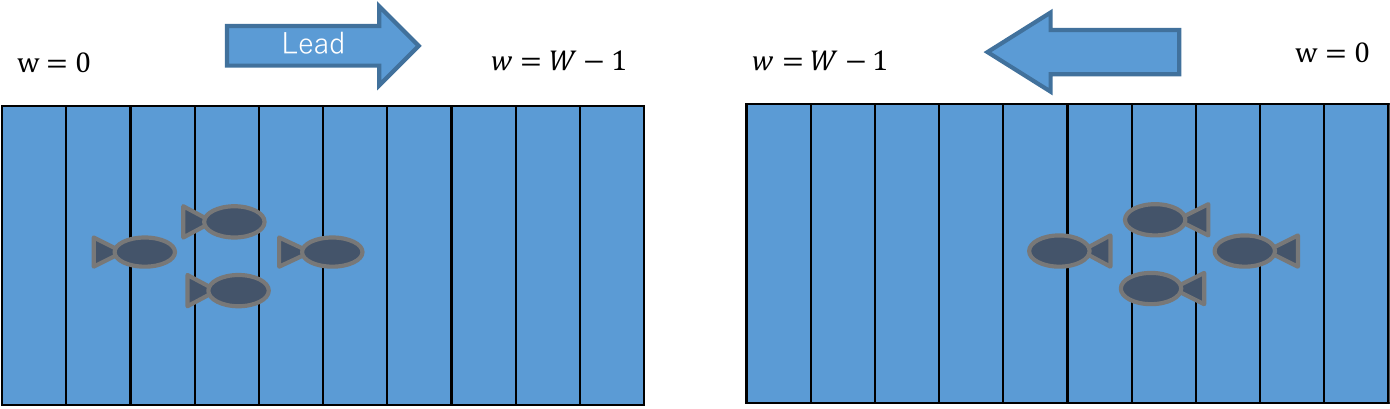}
    \caption{Definition of cells. When the number of divisions is $W$, the cell at the target end is defined as $W-1$, and the cell at the opposite end is defined as $0$.}
    \label{fig:cell}
\end{figure}

Before defining $\mathcal{S}$ and $\mathcal{A}$, we define regions called \textit{cells} within the viewport. Cells are defined by dividing the viewport horizontally into $W$ intervals. For a given number of divisions $W$, a cell $w \in \{0, 1, \dots, W-1\}$ is defined in the viewport coordinate system as:
\begin{align}
    \begin{dcases}
        \left\{ \begin{pmatrix} x \\ y \end{pmatrix} \,\middle|\, x \in \left[ \frac{w}{W}, \frac{w+1}{W} \right),\, y \in [0, 1] \right\} & \text{when guiding toward the right edge,} \\
        \left\{ \begin{pmatrix} x \\ y \end{pmatrix} \,\middle|\, x \in \left( \frac{W - w - 1}{W}, \frac{W - w}{W} \right],\, y \in [0, 1] \right\} & \text{when guiding toward the left edge.}
    \end{dcases}
\end{align}
This cell definition is illustrated in Fig.~\ref{fig:cell}.

The state is defined as a pair of $w_{\mathrm{real}}$ and $w_{\mathrm{virtual}}$, representing the cells containing the centroids of the real fish school and that of the virtual fish, respectively. Thus, the state space is:
\begin{align}
    \mathcal{S} = \{ (w_\mathrm{real}, w_\mathrm{virtual}) \mid w_\mathrm{real}, w_\mathrm{virtual} \in \{0, 1, 2, \dots, W-1\} \}.
\end{align}

Next, we define the actions. The movement of the virtual fish at each step is determined as follows:
\begin{description}
    \item[Step 1:] Determine the target point $\vec{x}_\mathrm{virtual\text{-}target}$.
        \begin{enumerate}
            \item Obtain the current centroid positions $\vec{x}_\mathrm{real}$ and $\vec{x}_\mathrm{virtual}$.
            \item Determine the cell $w_\mathrm{virtual}$ containing $\vec{x}_\mathrm{virtual}$.
            \item Select a cell displacement $\Delta w \in \{0, \pm 1, \pm 2, \dots, \pm \Delta w_\mathrm{max}\}$.
            \item Set the target cell $w_\mathrm{virtual\text{-}target}$:
                \begin{align}
                    \tilde{w}_\mathrm{virtual\text{-}target} &= w_\mathrm{virtual} + \Delta w \\
                    w_\mathrm{virtual\text{-}target} &= \begin{dcases}
                        0 & \text{if } \tilde{w}_\mathrm{virtual\text{-}target} < 0, \\
                        \tilde{w}_\mathrm{virtual\text{-}target} & \text{if } 0 \leq \tilde{w}_\mathrm{virtual\text{-}target} \leq W - 1, \\
                        W-1 & \text{if } W-1 < \tilde{w}_\mathrm{virtual\text{-}target}.
                    \end{dcases}
                \end{align}
            \item Using the $x$-coordinate center of $w_\mathrm{virtual\text{-}target}$ ($x_\mathrm{virtual\text{-}target}$) and the $y$-coordinate of the real fish centroid ($y_\mathrm{real}$), set the target point:
                \begin{align}
                    \vec{x}_\mathrm{virtual\text{-}target} = \begin{pmatrix} x_\mathrm{virtual\text{-}target} \\ y_\mathrm{real} \end{pmatrix}.
                \end{align}
        \end{enumerate}
    \item[Step 2:] Move toward the target point until the next step. The centroid follows a first-order lag motion:
        \begin{align}
            \label{eq:virtual-movement}
            \diffrac{\vec{x}_\mathrm{virtual}}{t} &= \frac{1}{\tau_\mathrm{virtual}}(\vec{x}_\mathrm{virtual\text{-}target} - \vec{x}_\mathrm{virtual}), \\
            \frac{1}{\tau_\mathrm{virtual}} &> 0 \quad (\text{constant}).
        \end{align}
\end{description}
Since the motion of the virtual fish is determined by $\Delta w$, we define $\Delta w$ as the action. Thus, the action space is:
\begin{align}
    \mathcal{A} = \{0, \pm 1, \pm 2, \dots, \pm \Delta w_\mathrm{max}\}.
\end{align}

Finally, the reward $r_n$ is defined based on the $x$-coordinate of the real fish school centroid $x_\mathrm{real}$ at step $n$:
\begin{align}
    r_n = \begin{dcases}
        2 \times (x_\mathrm{real} - 0.5) & \text{when guiding toward the right edge,} \\
        2 \times (0.5 - x_\mathrm{real}) & \text{when guiding toward the left edge.}
    \end{dcases}
\end{align}
Thus, $r_n \in [-1, +1]$, and $r_n$ approaches $+1$ as the centroid $\vec{x}_\mathrm{real}$ gets closer to the target edge.

\subsection{Reinforcement Learning via Q-Learning}

In this section, we describe the Q-learning algorithm, based again on Sutton and Barto~\cite{Sutton:1998}. The agent's actions are determined probabilistically according to a policy $\pi$. The probability distribution of action $a_n$ given state $s_n$ is:
\begin{align}
    a_n \sim \pi(\cdot | s_n),
\end{align}
where $\pi(a|s)$ represents the probability of choosing action $a$ in state $s$. The expected discounted return when following policy $\pi$ from state $s$ is defined as the state-value function:
\begin{align}
    V^\pi(s) = E_\pi \{ R_n | s_n = s \}.
\end{align}
The optimal policy $\pi^*$ is defined as the policy satisfying
\begin{align}
    \pi^* \in \argmax_\pi V^\pi(s)
\end{align}
for all $s \in \mathcal{S}$. The goal of RL is to acquire $\pi^*$. Given the optimal action-value function $Q^*(s, a) = \max_\pi Q^\pi(s, a)$, where $Q^\pi(s, a) = E_\pi \{ R_n | s_n = s, a_n = a \}$, the optimal policy can be obtained as:
\begin{align}
    \pi^*(a|s) = \begin{dcases}
        \frac{1}{|\argmax_{a'} Q^*(s, a')|} & \text{if } a \in \argmax_{a'} Q^*(s, a'), \\
        0 & \text{otherwise.}
    \end{dcases}
\end{align}
Based on the Bellman optimality equation,
\begin{align}
    Q^*(s, a) = E \left\{ r_{n+1} + \gamma \max_{a'} Q^*(s_{n+1}, a') \,\middle|\, s_n = s, a_n = a \right\},
\end{align}
the function $Q(s_n, a_n)$ can be updated iteratively to approximate $Q^*$ by making it converge toward $r_{n+1} + \gamma \max_{a'} Q(s_{n+1}, a')$. The learned policy $\pi$ corresponding to this approximated $Q$ is expressed as:
\begin{align}
    \label{eq:learned-policy}
    \pi(a|s) = \begin{dcases}
        \frac{1}{|\argmax_{a'} Q(s, a')|} & \text{if } a \in \argmax_{a'} Q(s, a'), \\
        0 & \text{otherwise.}
    \end{dcases}        
\end{align}
Q-learning implements this through the following procedure:
\begin{description}
    \item[Step 1:] Initialize the action-value function $Q$ and observe the initial state $s_0$.
    \item[Step 2:] Repeat for each step $n = 0, 1, \dots, N-1$:
        \begin{enumerate}
            \item Select action $a$ in state $s$ based on the policy $\pi_n$ derived from $Q$.
            \item Execute action $a$ and observe reward $r$ and next state $s'$.
            \item Update $Q$ according to:
                \begin{align}
                    Q(s, a) &\leftarrow Q(s, a) + \alpha \left[ r + \gamma \max_{a'} Q(s', a') - Q(s, a) \right], \\
                    \alpha &\in (0, 1) \quad (\text{constant, called the ``step size''}).
                \end{align}
            \item $s \leftarrow s'$
        \end{enumerate}
\end{description}
Regarding $\pi_n$, it is effective to perform \textit{exploration} (trying various actions) in the early stages of learning and follow the best-known policy in the later stages. Thus, we use an $\epsilon$-greedy algorithm where $\epsilon$ is decayed at each step:
\begin{align}
    \epsilon_n &= 1 - \frac{n}{N-1} \\
    \pi_n(a|s) &= \begin{dcases}
        \frac{1 - \epsilon_n}{|\argmax_{a'} Q(s, a')|} + \frac{\epsilon_n}{|\mathcal{A}|} & \text{if } a \in \argmax_{a'} Q(s, a'), \\
        \frac{\epsilon_n}{|\mathcal{A}|} & \text{otherwise.}
    \end{dcases}
\end{align}

\section{Preliminary Evaluation in a Simulation Environment}
\label{sec:simulation}
When interacting with a fish school using virtual fish, the frequency with which the real fish react to the virtual fish is expected to be limited. To investigate whether an effective policy can be acquired via RL under such conditions, we perform RL in a simulation environment that mimics the movement of real fish.

Furthermore, as will be discussed in Section~\ref{sec:realworld}, learning an effective policy in an actual interaction system is expected to be extremely time-consuming. By using the action-value function $Q$ obtained in the simulation environment, where each time step can be processed much faster than in the real environment, as an initial value, we expect to reduce the learning time required in the real-world experiments.

\subsection{Modeling the Movement of Real Fish}
\label{sec:behavior-model}
We simulate the behavior of real fish using an action model in which the decision to react to the virtual fish is determined probabilistically. By varying this probability, we examine whether RL can acquire effective policies even when real fish infrequently react to virtual fish.

Based on observations of the fish during housing, the following behavioral characteristics were identified:
\begin{itemize}
    \item They show minimal reaction to other individuals that are far away.
    \item When they do react to another individual, they often begin swimming toward that individual.
    \item Most of their movements can be viewed as a repetition of linear motions lasting for a few seconds.
\end{itemize}
We developed a model for real fish that reflects these properties.

For simplicity, we assume that the real fish always act as a collective school and simulate the motion of their centroid. The following parameters are held constant within each simulation:
\begin{itemize}
    \item $\Delta{t}_\mathrm{max} > 0$: Maximum duration before the real fish switch their target point.
    \item $\theta > 0$: Maximum distance at which real fish react to the virtual fish.
    \item $p \in [0,1]$: Probability that the real fish ignore the virtual fish even when they are sufficiently close.
    \item $\delta{x}_\mathrm{max} > 0$, $\delta{y}_\mathrm{max} > 0$: Maximum displacement of the target point when the virtual fish are ignored.
    \item $1/\tau_\mathrm{real} > 0$: Inverse of the time constant for the first-order lag system of the real fish.
\end{itemize}

Let $\vec{x}_\mathrm{real} = \begin{pmatrix}x_\mathrm{real}\\y_\mathrm{real}\end{pmatrix}$ and $\vec{x}_\mathrm{virtual}$ represent the centroid positions of the real fish school and the virtual fish, respectively. The motion of $\vec{x}_\mathrm{real}$ is modeled as a series of movements toward a switching target point $\vec{x}_\mathrm{real\text{-}target}$, each of which is governed by a first-order lag system, simulated as follows:

\begin{description}
    \item[Step 1:] Randomly select a duration $\Delta{t} \in (0,\Delta{t}_\mathrm{max}]$ for which the fish move linearly.
    \item[Step 2:] Determine the target point $\vec{x}_\mathrm{real\text{-}target}$ as follows:
        \begin{itemize}
            \item If $\norm{\vec{x}_\mathrm{virtual} - \vec{x}_\mathrm{real}} \leq \theta$:
                \begin{itemize}
                    \item With probability $1-p$ (reaction case), set:
                        \begin{align}
                            \vec{x}_\mathrm{real\text{-}target} = \vec{x}_\mathrm{virtual}.
                        \end{align}
                    \item With probability $p$ (ignore case), select random relative displacements $\delta{x} \in [-\delta{x}_\mathrm{max}, +\delta{x}_\mathrm{max}]$ and $\delta{y} \in [-\delta{y}_\mathrm{max}, +\delta{y}_\mathrm{max}]$, and set:
                        \begin{align}
                            \label{eq:x-real-target-sim}
                            \vec{x}_\mathrm{real\text{-}target} = \vec{x}_\mathrm{real} +
                                \begin{pmatrix}
                                    \delta{x}\\
                                    \delta{y}
                                \end{pmatrix}.
                        \end{align}
                \end{itemize}
            \item Otherwise (out of range):
                Determine $\vec{x}_\mathrm{real\text{-}target}$ using random displacements as in Eq.~\eqref{eq:x-real-target-sim}.
        \end{itemize}

    \item[Step 3:] During the interval $\Delta{t}$, move toward $\vec{x}_\mathrm{real\text{-}target}$ based on
        \begin{align}
            \diffrac{\vec{x}_\mathrm{real}}{t}
                = \frac{1}{\tau_\mathrm{real}}(\vec{x}_\mathrm{real\text{-}target} - \vec{x}_\mathrm{real}).
        \end{align}
    \item[Step 4:] Repeat Steps 1 through 3.
\end{description}

\subsection{Evaluation of Robustness Against Stochastic Behavior}

\subsubsection{Simulation Evaluation Conditions}
\label{sec:simulation-eval-conditions}

We evaluate whether RL can acquire effective movement policies even in scenarios where real individuals infrequently respond to the virtual stimulus. To this end, we conduct RL for each ignoring probability $p$ defined in the behavioral model in Section~\ref{sec:behavior-model}. We then investigate whether the resulting policy approximates the optimal policy by analyzing the rewards obtained when following this policy.

For each ignoring probability $p = 0, 0.3, 0.6, 0.9$, Q-learning is performed with the number of training steps $N$ varied across $\{10^2, 10^3, 10^4, 10^5, 10^6, 10^7\}$. Other parameter settings are summarized in Table~\ref{tab:simulation-param}. After training, the agent follows the learned policy $\pi$ (as defined in Eq.~\eqref{eq:learned-policy}) for $M$ steps to calculate the average reward $R$:
\begin{equation}
    R = \frac{1}{M}\sum_{n=0}^{M} r_n,
\end{equation}
where $M = 9000$. Since both the learned policy and the resulting reward $R$ are inherently stochastic, we expect the evaluation $R$ to vary across trials. Therefore, we repeat the training and evaluation process 10 times for each combination of $p$ and $N$, and calculate the mean of $R$ across these trials.

\begin{table}[tbp]
    \centering
    \caption{Simulation parameters. $\theta$ and $\delta{x}_\mathrm{max}$ are relative to the viewport width, and $\delta{y}_\mathrm{max}$ is to the viewport height.}
    \label{tab:simulation-param}
    \begin{tabular}{|c|l|c|}
        \hline
        Parameter & Description & Value\\
        \hline
        \hline
        $\gamma$ & Discount factor (Q-learning) & 0.9\\
        $\alpha$ & Step size (Learning rate) (Q-learning) & 0.1\\
        \hline
        $W$ & Number of cells & 10\\
        $\Delta{w}_\mathrm{max}$ & Max cell displacement per step (virtual fish) & 2\\
        $1/\tau_\mathrm{virtual}$ & Inverse time constant (virtual fish) & \SI{3}{s^{-1}}\\
        \hline
        $\Delta{t}_\mathrm{max}$ & Max duration for target switching (real fish) & \SI{3}{s}\\
        $\theta$ & Max reaction distance (real fish) & 0.3\\
        $\delta{x}_\mathrm{max}$ & Max $x$-displacement during ignore cases (real fish) & 0.2\\
        $\delta{y}_\mathrm{max}$ & Max $y$-displacement during ignore cases (real fish)& 0.2\\
        $1/\tau_\mathrm{real}$ & Inverse time constant (real fish) & \SI{1}{s^{-1}} \\
        \hline
    \end{tabular}
    \vspace{3mm}
\end{table}

Note that the reward $R$ obtained under an optimal policy varies depending on the ignoring probability $p$. To provide a benchmark for evaluating the learned policies, we define a heuristic policy $\hat{\pi}$ that is expected to yield near-optimal rewards. In this simulation, the maximum interaction distance $\theta$ (the range within which real fish react to the virtual fish) is set to 0.3 times the viewport width, and the number of cell divisions is $W = 10$. Consequently, a policy $\hat{\pi}$ that targets a position three cells ahead of the real fish's current cell $w_{\mathrm{real}}$ (i.e., $w_{\mathrm{real}} + 3$) is considered near-optimal. Specifically, given a state $(w_{\mathrm{real}}, w_{\mathrm{virtual}})$, the baseline policy $\hat{\pi}$ is defined as follows:
\begin{align}
    \Delta\tilde{w} &=  w_{\mathrm{real}} + 3 - w_{\mathrm{virtual}}. \\
    \Delta{w} &= \begin{dcases}
        -\Delta{w}_{\mathrm{max}} & \text{if } \Delta\tilde{w} < -\Delta{w}_{\mathrm{max}}, \\
        \Delta\tilde{w} & \text{if } -\Delta{w}_{\mathrm{max}} \leq \Delta\tilde{w} \leq \Delta{w}_{\mathrm{max}}, \\
        +\Delta{w}_{\mathrm{max}} & \text{if } \Delta{w}_{\mathrm{max}} < \Delta\tilde{w}.
    \end{dcases} \\
    \label{eq:sim-baseline}
    \hat{\pi}(a|s) &= \begin{dcases}
        1 & \text{if } a = \Delta{w}, \\
        0 & \text{otherwise.}
    \end{dcases}
\end{align}
Here, the maximum displacement of the virtual fish's cell per step is set to $\Delta{w}_{\mathrm{max}} = 2$, consistent with the constraints of the RL agent.

We denote the reward obtained by this baseline policy $\hat{\pi}$ over $M = 9000$ steps as $\hat{R}$. Since $\hat{R}$ is also expected to vary across trials, we calculate $\hat{R}$ 10 times for each $p$ and take the mean. If the mean reward $R$ of a learned policy is sufficiently close to the mean benchmark $\hat{R}$, we consider the acquired policy to be near-optimal.

\subsubsection{Results and Discussion of Simulation Evaluation}
\label{sec:simulation-results}

\begin{figure}[tbp]
    \centering
    \includegraphics[width=\linewidth]{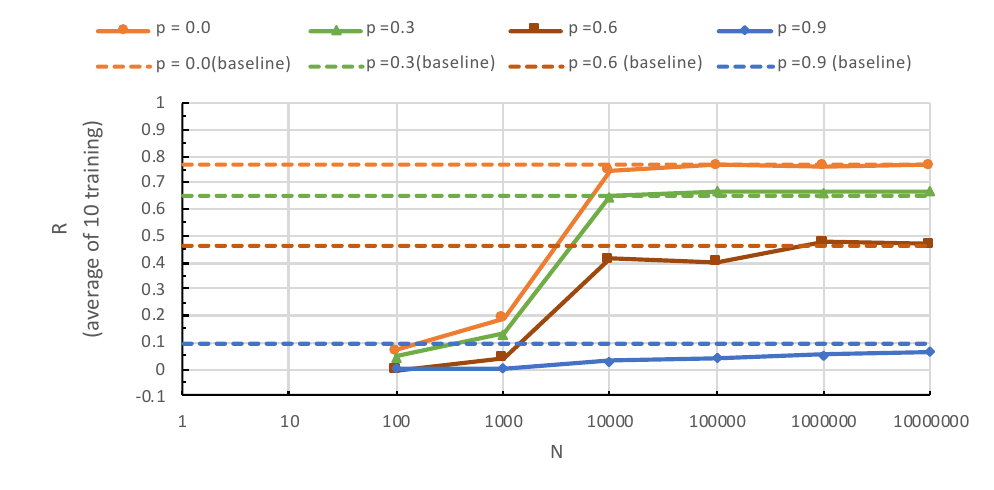}
    \caption{Average reward $R$ in simulation. $p$ represents the probability of the simulated real fish ignoring the virtual fish. The baseline denotes the reward obtained using $\hat{\pi}$ as defined in Eq.~\eqref{eq:sim-baseline}.}
    \label{fig:simulation-eval}
\end{figure}

The results of the evaluation described in Section~\ref{sec:simulation-eval-conditions} are shown in Fig.~\ref{fig:simulation-eval}. Regardless of the ignoring probability $p$, it can be observed that as the number of training steps $N$ increases, the average reward $R$ obtained under the learned policy approaches the reward $\hat{R}$ obtained under the baseline policy $\hat{\pi}$.

\begin{figure}[tbp]
    \centering
    \includegraphics[width=\linewidth]{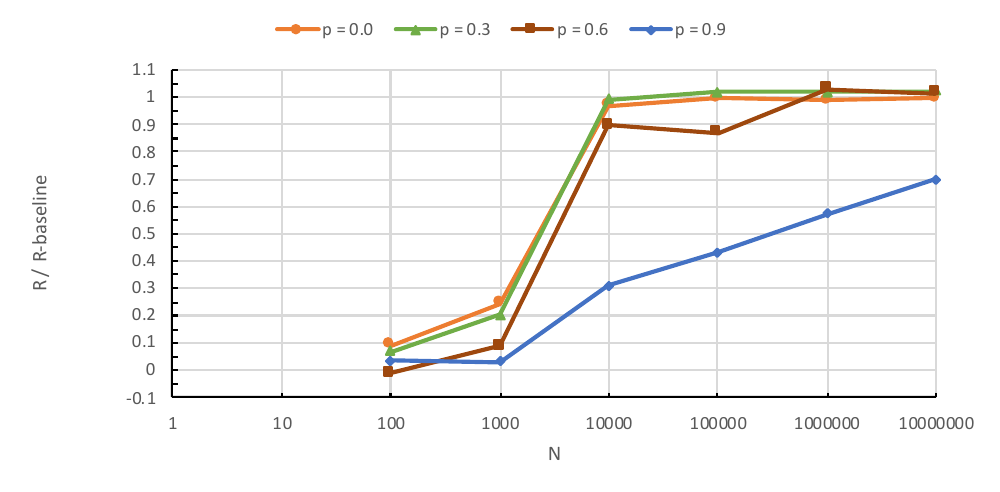}
    \caption{Ratio of the average reward of the learned policy to that of the baseline policy in simulation. $p$ represents the probability of the simulated real fish ignoring the virtual fish.}
    \label{fig:simulation-eval-ratio}
\end{figure}

Figure~\ref{fig:simulation-eval-ratio} shows the ratio of the mean reward $R$ under the learned policy to the mean reward $\hat{R}$ under the baseline policy. For ignoring probabilities $p = 0, 0.3,$ and $0.6$, the ratio reaches approximately 1.0 when the number of training steps $N$ is sufficiently large. This suggests that the learned policies in these cases are nearly optimal. Furthermore, even in the case of $p = 0.9$, where the real fish ignore the virtual stimulus with high frequency, the ratio reaches approximately 0.7 at $N = 10^7$, indicating that learning progresses to a certain extent even under such challenging conditions.

\section{Reinforcement Learning in a Real-World Environment}
\label{sec:realworld}

\subsection{Acquisition of Centroid Positions for Real and Virtual Fish}
The centroid position of the real fish is determined as follows. For an RGB color image, let $\vec{p}_{u,v} = \begin{pmatrix} r_{u,v} \\ g_{u,v} \\ b_{u,v} \end{pmatrix}$ represent the pixel at column $u$ and row $v$. The entire image is denoted as $\setEx{\vec{p}_{u,v}}$, where $r_{u,v}$, $g_{u,v}$, and $b_{u,v}$ are the red, green, and blue values at position $(u,v)$. For a grayscale image, the pixel value is denoted as $p_{u,v}$, and the image as $\setEx{p_{u,v}}$.

\begin{description}
    \item[Step 1:]
        A binary image is obtained using background subtraction. Given a pre-recorded background image $\setEx{\tilde{\vec{p}}_{u,v}}$ and a threshold $\theta$, we compute for each $(u,v)$:
        \begin{align}
            p'_{u,v} = \begin{dcases}
                1 & \text{if } \norm{\vec{p}_{u,v} - \tilde{\vec{p}}_{u,v}} > \theta, \\
                0 & \text{otherwise.}
            \end{dcases} 
        \end{align}
        The resulting binary image $\setEx{p'_{u,v}}$ is expected to have a value of 1 where real or virtual fish are present, and 0 elsewhere.
        
    \item[Step 2:]
        As shown in Fig.~\ref{fig:observed-image}, real fish appear as dark silhouettes due to the display's backlighting. This color difference is used to distinguish between real and virtual fish. Using a decision function $f$ obtained via a Support Vector Machine (SVM), we compute:
        \begin{align}
            p''_{u,v} = \begin{dcases}
                1 & \text{if } p'_{u,v} = 1 \text{ and } f(\vec{p}_{u,v}) > 0, \\
                0 & \text{otherwise.}
            \end{dcases}
        \end{align}
        The SVM is pre-trained on a dataset where $f(\vec{p})$ is $+1$ for real fish pixels and $-1$ for virtual fish pixels.

    \item[Step 3:]
        The binary image $\setEx{p''_{u,v}}$ is processed with an opening operation (an equal number of erosions and dilations) to obtain $\setEx{p'''_{u,v}}$. This ensures that pixels with a value of 1 correspond only to the real fish.

    \item[Step 4:]
        The centroid $\tilde{\vec{x}}_\mathrm{real}$ of the pixels with value 1 in $\setEx{p'''_{u,v}}$ is calculated as:
        \begin{align}
            \tilde{\vec{x}}_\mathrm{real} = \begin{pmatrix}
                \frac{m_{1,0}}{m_{0,0}}\\
                \frac{m_{0,1}}{m_{0,0}}
            \end{pmatrix}
            \quad \text{where} \quad
            m_{i,j} = \sum_{u,v} u^i v^j p'''_{u,v}.
        \end{align}
        Finally, $\tilde{\vec{x}}_\mathrm{real}$ is converted to viewport coordinates $\vec{x}_\mathrm{real}$ using Eqs.~\eqref{eq:camera-to-viewport-x} and~\eqref{eq:camera-to-viewport-y}.
\end{description}
Step 3 is a procedure designed to eliminate misclassifications originating from the SVM in Step 2. Specifically, classification errors occasionally caused certain subregions to be marked as containing a real fish ($p''_{u,v} = 1$) even when no fish was present. By implementing Step 3, we successfully removed most of these false positives.

Regarding the position of the virtual fish centroid, we set the time constant $\tau_{\mathrm{imitated}}$ of the first-order lag system in Eq.~\eqref{eq:virtual-movement} to be sufficiently small so that the movement speed is high enough. This allows us to assume that the virtual fish reach sufficiently close to their target point $\vec{x}_{\mathrm{virtual\text{-}target}}$ within a single time step. Consequently, the position of the virtual fish centroid at any given time is treated as the target point selected in the preceding state.

\subsection{Evaluation of the Learned Policy}
\label{sec:evalulation-learned-policy}

We investigate whether RL can yield an effective movement policy for virtual fish to guide fish schools in a real-world environment. Our analysis is based on the behavior of both real and virtual individuals under the learned policy. Since training in a real-world environment is expected to be extremely time-consuming, we initialize the action-value function $Q$ using the values obtained from simulation (with an ignoring probability $p = 0.6$ and $N = 10^6$ steps) to shorten the required training duration.

\begin{table}[tbp]
    \centering
    \caption{Parameters for RL and control in the real-world environment.}
    \label{tab:real-env-param}
    \begin{tabular}{|c|l|c|}
        \hline
        Parameter & Description & Value \\
        \hline
        \hline
        $\gamma$ & Discount factor (Q-learning) & 0.9 \\
        $\alpha$ & Step size (Learning rate) (Q-learning) & 0.1 \\
        \hline
        $W$ & Number of cell divisions & 10 \\
        $\Delta w_\mathrm{max}$ & Max cell displacement per step (virtual fish) & 2 \\
        $1/\tau_\mathrm{virtual}$ & Inverse time constant (virtual fish) & \SI{1}{s^{-1}} \\
        \hline
    \end{tabular}
    \vspace{3mm}
\end{table}

Reinforcement learning was conducted for $N = 1800$ steps. The parameter settings are summarized in Table~\ref{tab:real-env-param}. Three randomly selected real fish were used for the experiment. Subsequently, 900 steps of control were performed on the same individuals following the learned policy (hereafter referred to as the ``proposed method''). We verify whether the proposed method successfully guides the real individuals toward the target direction and determine if it is more effective than a simple, manually defined policy. We therefore investigated the behavior of real individuals for 900 steps under the following baselines:
\begin{description}
    \item[Baseline 1 (stay at edge):] The virtual fish move toward the target direction and stays at the edge of the viewport once it reaches it.
    \item[Baseline 2 (w/o stimulus):] No virtual fish are displayed.
\end{description}
Here, the policy for Baseline 1 was defined as follows, with the maximum displacement per step set to $\Delta{w}_{\mathrm{max}} = 2$ to remain consistent with the proposed method:
\begin{equation}
\pi(a|s) = \begin{dcases}
1 & \text{if } a = +\Delta{w}_{\mathrm{max}}, \\
0 & \text{otherwise.}
\end{dcases}
\end{equation}
The reactions of real fish to the virtual fish are expected to be influenced by the preceding movements of the virtual stimuli. Therefore, to avoid carry-over effects from previous trials, a rest period of at least 10 minutes was provided between the training phase, the evaluation of the proposed method, and each baseline experiment.

\begin{figure}[tbp]
    \centering
    \begin{minipage}{\linewidth}
        \centering
        \includegraphics[width=\linewidth]{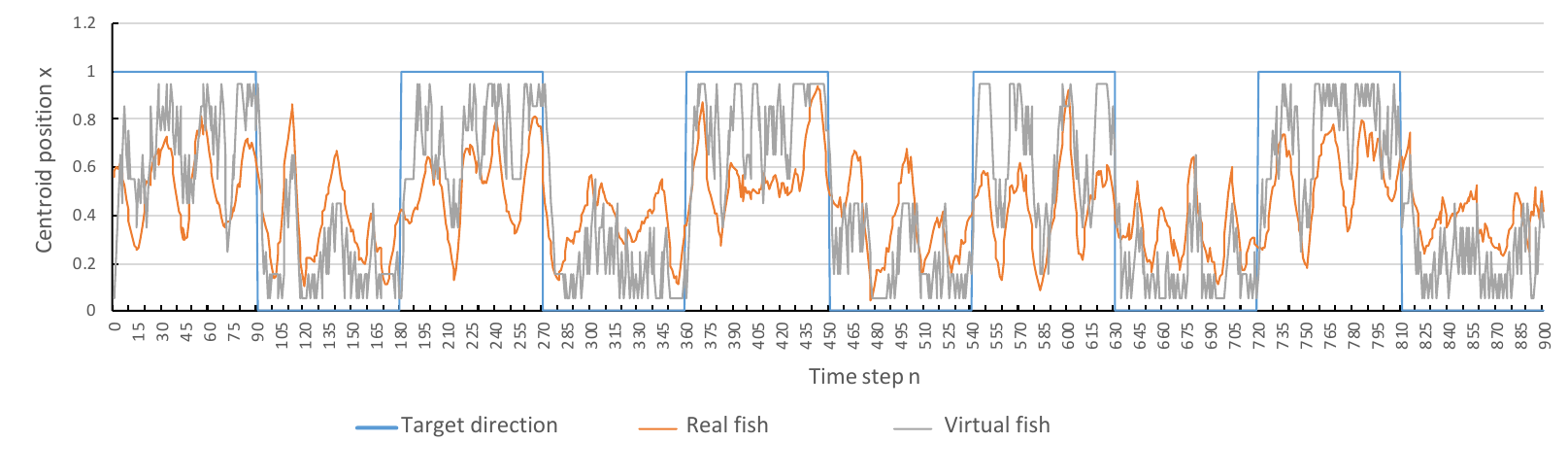}
        \subcaption{Proposed Method}
    \end{minipage} \\ \vspace{1em}
    \begin{minipage}{\linewidth}
        \centering
        \includegraphics[width=\linewidth]{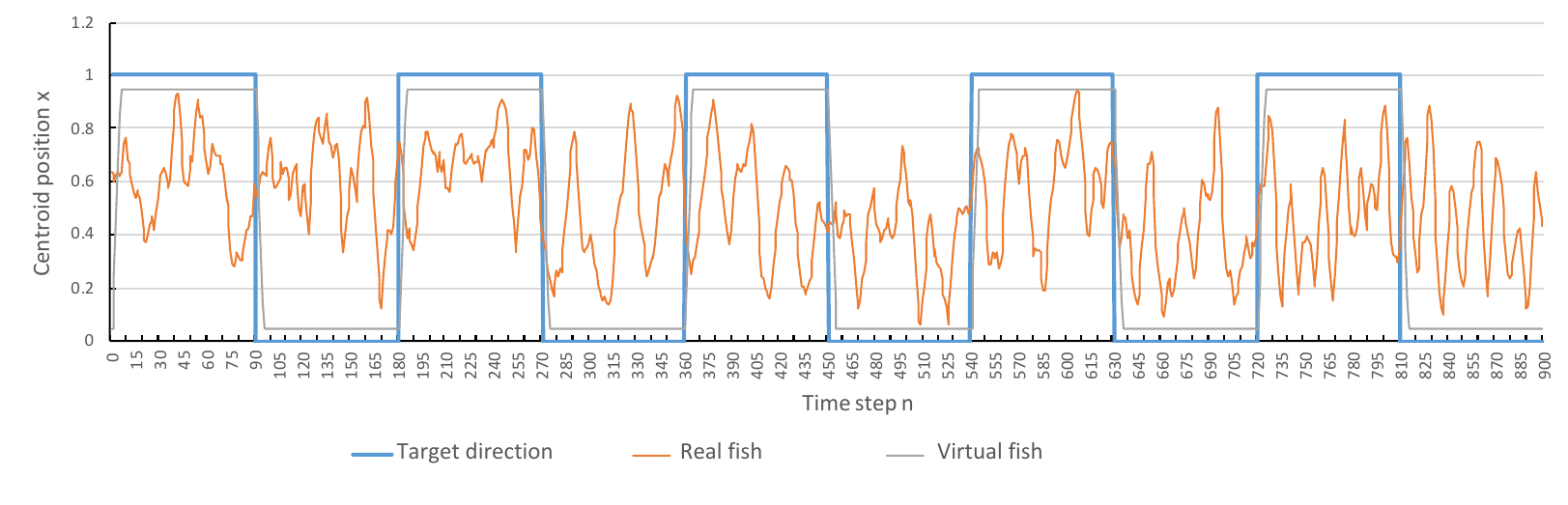}			
        \subcaption{Baseline 1 (stay at edge)}
    \end{minipage} \\ \vspace{1em}
    \begin{minipage}{\linewidth}
        \centering
        \includegraphics[width=\linewidth]{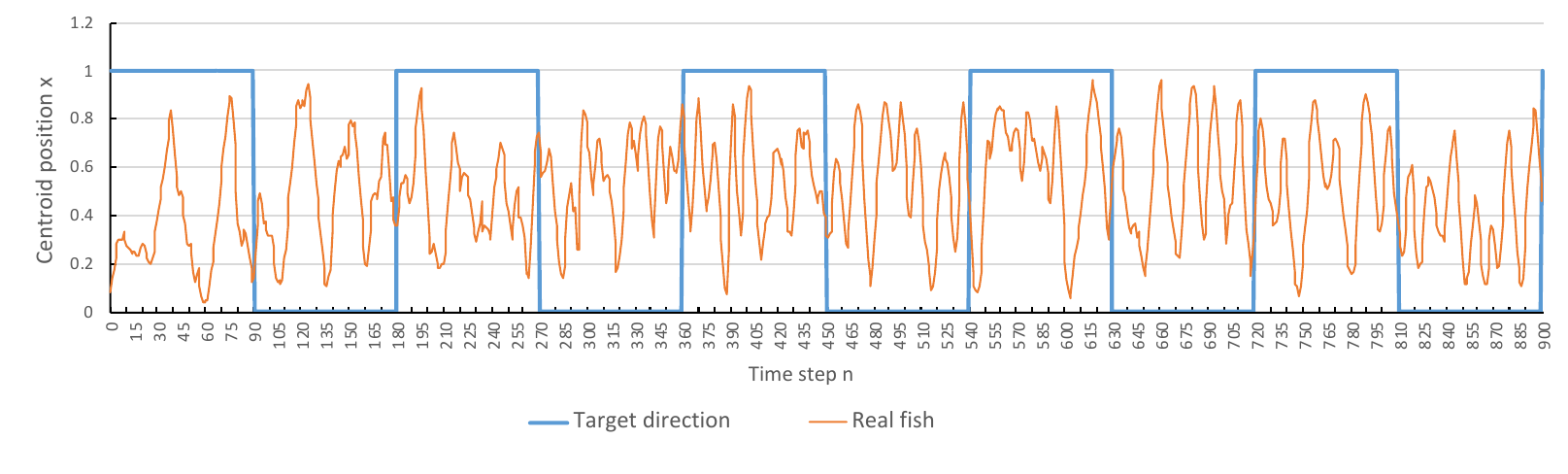}
        \subcaption{Baseline 2 (w/o stimulus)}				
    \end{minipage}
    \caption{Time series of the $x$-coordinate of the real fish centroid during real-world control.}
    \label{fig:real-eval-timeline}
\end{figure}

The centroids of both real and virtual fish were recorded at each time step and are summarized in Fig.~\ref{fig:real-eval-timeline}. In the proposed method, the virtual fish move within the area between the target position and the area near the real fish, appearing to guide the real individuals toward the goal.

\begin{figure}[tbp]
    \centering
    \begin{minipage}{\linewidth}
        \centering
        \includegraphics{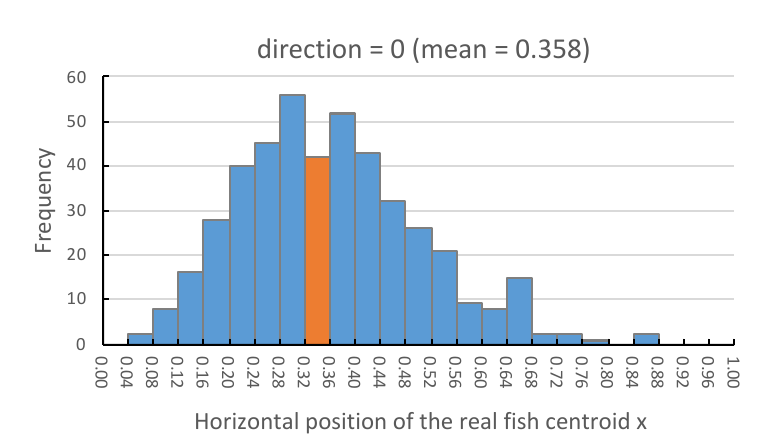}
        \subcaption{Guiding toward the left ($x=0$)}
    \end{minipage} \\ \vspace{0.5em}
    \begin{minipage}{\linewidth}
        \centering
        \includegraphics{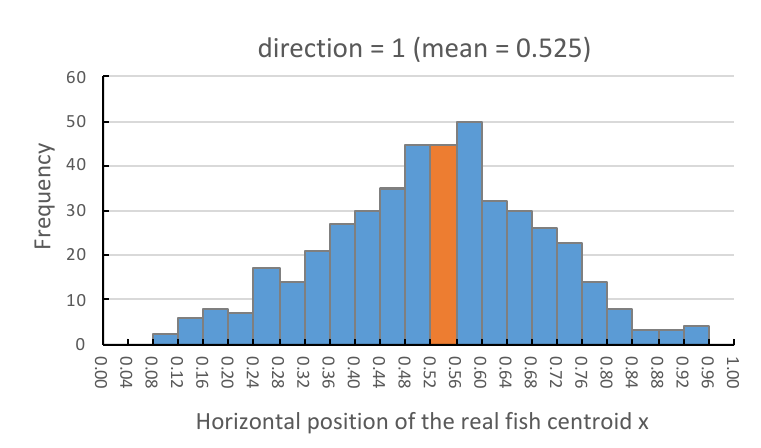}
        \subcaption{Guiding toward the right ($x=1$)}
    \end{minipage}
    \caption{Histograms of real fish $x$-coordinates (Proposed Method). Orange bins indicate the mean value's interval.}
    \label{fig:real-eval-hist-rl}
\end{figure}

\begin{figure}[tbp]
    \centering
    \begin{minipage}{\linewidth}
        \centering
        \includegraphics{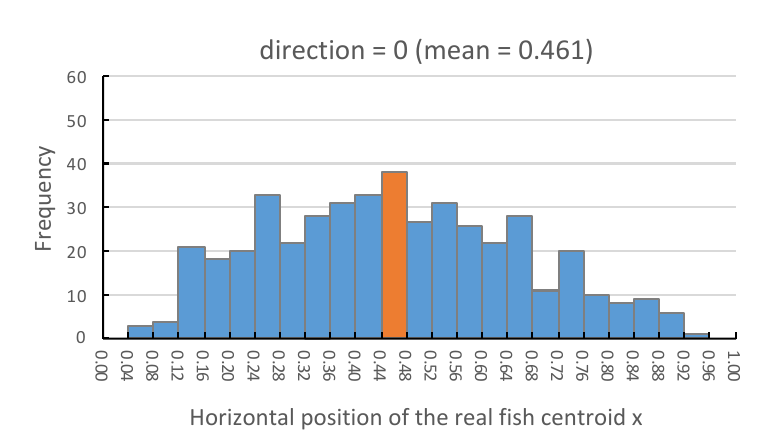}
        \subcaption{Guiding toward the left ($x=0$)}
    \end{minipage} \\ \vspace{0.5em}
    \begin{minipage}{\linewidth}
        \centering
        \includegraphics{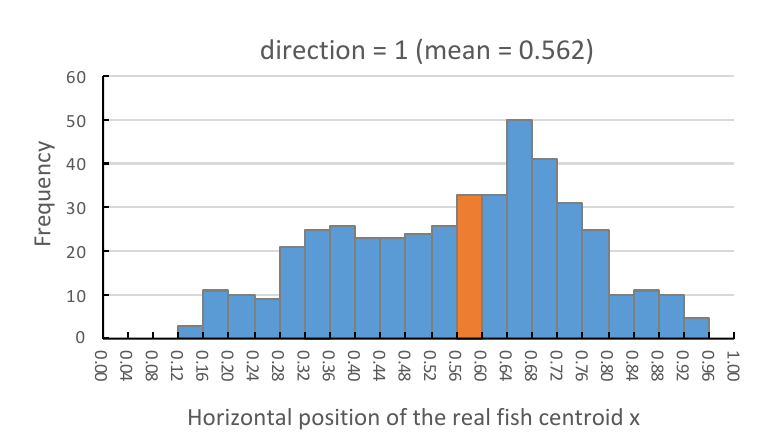}
        \subcaption{Guiding toward the right ($x=1$)}
    \end{minipage}
    \caption{Histograms of real fish $x$-coordinates (Baseline 1: stay at edge).}
    \label{fig:real-eval-hist-baseline}
\end{figure}

\begin{figure}[tbp]
    \centering
    \begin{minipage}{\linewidth}
        \centering
        \includegraphics{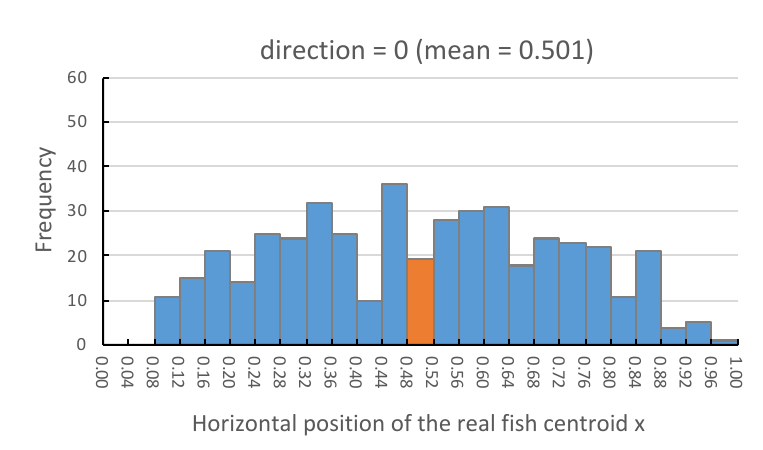}
        \subcaption{Guiding toward the left ($x=0$)}
    \end{minipage} \\ \vspace{0.5em}
    \begin{minipage}{\linewidth}
        \centering
        \includegraphics{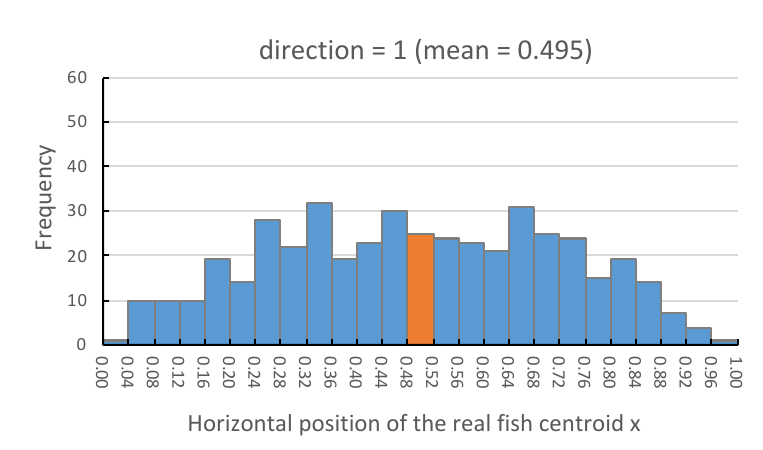}
        \subcaption{Guiding toward the right ($x=1$)}
    \end{minipage}
    \caption{Histograms of real fish $x$-coordinates (Baseline 2: w/o stimulus).}
    \label{fig:real-eval-hist-empty}
\end{figure}

We then generated histograms of the positions of real individuals (centroids) for each target guidance direction. Figures~\ref{fig:real-eval-hist-rl}, \ref{fig:real-eval-hist-baseline}, and \ref{fig:real-eval-hist-empty} show the results for the Proposed Method, Baseline 1, and Baseline 2, respectively. Table~\ref{tab:mean} summarizes the mean positions and the differences between them for each target direction. A comparison reveals that the magnitude of the difference in mean positions follows the order: (a) the Proposed Method, (b) Baseline 1 (stay at edge), and (c) Baseline 2 (w/o stimulus). The histograms also appear to be sharper in the same order.

\begin{table}[tbp]
    \centering
    \caption{Mean real fish $x$-coordinates for each guidance direction and the difference between them for each control method.}
    \label{tab:mean}
    \begin{tabular}{|l|S[table-format=1.3]|S[table-format=1.3]|S[table-format=-1.3]|}
        \hline
        Condition & {(a) Target: Left} & {(b) Target: Right} & {Difference (b-a)} \\
        \hline
        \hline
        Proposed Method & 0.358 & 0.525 & 0.166 \\
        Baseline 1 (stay at edge) & 0.461 & 0.562 & 0.101 \\
        Baseline 2 (w/o stimulus) & 0.501 & 0.495 & -0.006 \\
        \hline
    \end{tabular}
    \vspace{5mm}
\end{table}

We conducted Welch's $t$-test (two-tailed) at a significance level of \SI{5}{\%} to determine whether the target direction resulted in a significant difference in the mean positions of real individuals. The sample was defined as
\begin{align}
\{x_{\mathrm{real}}(t_n) \mid \text{guided to } x=0 \text{ (or } x=1) \text{ at step } n \},
\end{align}
assuming an approximately normal distribution. The resulting $p$-values are shown in Table~\ref{tab:ttest}. These values indicate that while there was no significant difference in (c) Baseline 2, significant differences were observed in both (a) the Proposed Method and (b) Baseline 1. Thus, it can be concluded that the real individuals were significantly guided in the intended directions in both (a) and (b).

\begin{table}[tbp]
    \centering
    \caption{Results of the significance tests for the difference in the mean $x$-coordinates of real fish based on the target guidance direction. A two-tailed Welch's $t$-test was performed for each virtual fish control method to evaluate the statistical significance of the difference.}
    \label{tab:ttest}
    \begin{tabular}{|l|S[table-format=1.2e-2]|}
        \hline
        Condition & {$p$-value} \\
        \hline
        \hline
        Proposed Method & 1.46e-51 \\
        Baseline 1 (stay at edge) & 8.48e-15 \\
        Baseline 2 (w/o stimulus) & 5.80e-1 \\
        \hline			
    \end{tabular}
    \vspace{3mm}
\end{table}

\begin{figure}[tbp]
    \centering
    \includegraphics[width=0.96\linewidth]{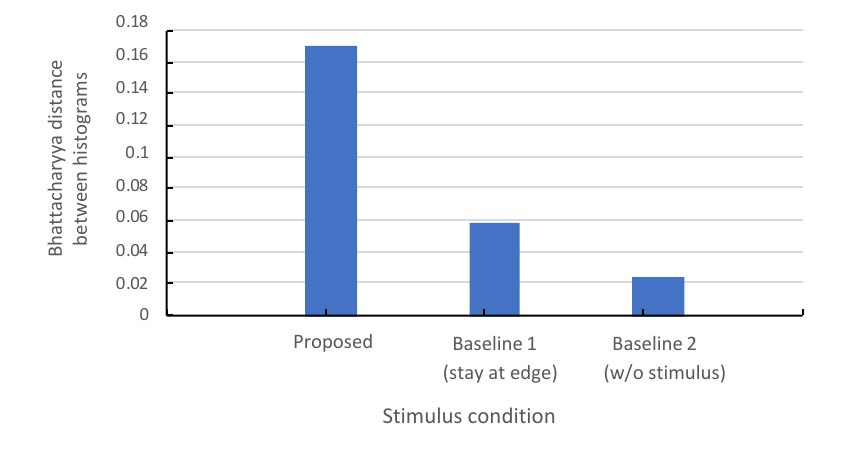}
    \caption{Distances between the $x$-coordinate distributions of real fish for leftward and rightward guidance. For each pair of histograms in Figs.~\ref{fig:real-eval-hist-rl} to \ref{fig:real-eval-hist-empty}, the effectiveness of the guidance was quantified by calculating the Bhattacharyya distance between the distributions resulting from the two target directions.}
    \label{fig:real-eval-bd}
\end{figure}

Furthermore, we evaluated the degree of difference between the histograms of the $x$-coordinates for each target direction (Figs.~\ref{fig:real-eval-hist-rl} to \ref{fig:real-eval-hist-empty}) using the Bhattacharyya distance. As shown in Fig.~\ref{fig:real-eval-bd}, the results confirm that the difference in the distribution of the real individuals' positions follows the order: (a) the Proposed Method, (b) Baseline 1 (stay at edge), and (c) Baseline 2 (w/o stimulus). Consequently, it is evident that the Proposed Method (a) guided the real individuals more prominently than Baseline 1 (b).

\section{Conclusion}
\label{sec:conclusion}

In this study, we investigated whether effective movement policies for virtual fish to guide fish schools could be acquired through reinforcement learning (RL).

First, there was a concern regarding whether RL could yield effective policies in scenarios where the frequency of real fish responding to virtual stimuli is not necessarily high. To address this, we established a behavioral model in which real individuals stochastically decide whether to follow or ignore virtual fish when they are sufficiently close. Our simulation results confirmed that reasonably effective policies could be obtained even when the probability of real individuals ignoring the virtual stimuli was high.

Second, we verified the feasibility of fish guidance in a real-world environment using the learned policy. The results demonstrated that when the virtual fish moved based on the learned policy, the mean positions of the real fish (centroids) differed significantly depending on the target guidance direction. This confirms that the real individuals were successfully guided in the intended directions. Furthermore, we compared the learned policy with a manual baseline policy where the virtual fish simply move to and stay at the edge of the viewport in the target direction. This comparison revealed that the distribution of the real fish's positions changed more prominently under the control of the learned policy than under the simple manual policy. These findings confirm that RL can acquire a more effective policy than a simple, human-defined one.

While this study demonstrated the potential of RL-based fish school control, several issues remain for future investigation.

\begin{description}
\item[Validation of image-based stimuli as virtual fish:]
In this study, we used textures of real individuals for the visual stimuli. However, it remains unclear whether these stimuli are recognized by real fish as actual members of their school. To clarify this, it is necessary to investigate whether behavioral responses differ when presenting abstract stimuli, such as striped patterns or random dots, compared to the texture-mapped virtual fish used in our experiments.

\item[Investigation of the number of individuals]
In this study, the numbers of both virtual and real fish were fixed throughout the experiments. However, toward the future goal of guiding large-scale schools with a small number of virtual or robotic agents, it is necessary to examine how the interaction dynamics change when the respective numbers of real and virtual individuals are varied.

\item[Examination of state representation in RL]
The state defined in this study was based on the respective centroids of the real fish school and the virtual fish group. However, even if the school is at the same position, the optimal action is expected to differ depending on whether the school is moving toward or away from the target direction. Therefore, potential improvements include incorporating information regarding the school's velocity into the state representation.

Furthermore, for the future goal of interacting with only a subset of individuals within a large-scale school, it will be necessary to target specific individuals that exert a significant influence on the collective motion. Defining the state in such scenarios would require incorporating more microscopic information, such as the spatial relationships among individuals and the status or strength of inter-individual interactions.

On the other hand, there is a trade-off where the training duration for RL increases as the size of the state space grows. Therefore, it is necessary to devise methods to incorporate the aforementioned information while effectively managing the number of elements in the state set.

\item[Examination of action representation in RL]
In the experiments conducted in this study, the movement of the four virtual fish was designed such that they moved as a single coordinated unit while maintaining their relative positions. However, to achieve more effective interactions with the same number of agents, it is expected that having each virtual fish move independently to interact with different individuals within the school would be more efficient.

On the other hand, similar to the challenges faced with the state space, it is necessary to devise methods that enable independent movement for each virtual fish while simultaneously limiting the size of the action space to a manageable level.
\end{description}

\section*{Ethical Approval}
All animal experiments were conducted with the approval of the Graduate School of Informatics, Kyoto University (Approval No. Inf-K17007, March 21, 2017).

\end{document}